\pdfoutput = 1
\RequirePackage{fix-cm}
\documentclass[twocolumn]{svjour3}          
\smartqed  
\usepackage{graphicx}
\usepackage{mathptmx} 
\usepackage{amsmath}   
\usepackage{adjustbox}
\usepackage{natbib}
\usepackage{caption}

\begin{document}
\emergencystretch 3em
\bibliographystyle{spbasic}

\title{A matching based clustering algorithm for categorical data}


\author{Ruben A. Gevorgyan  \and  Yenok B. Hakobyan}

\institute{Ruben A Gevorgyan \at
             Faculty of Economics and Management, Yerevan State University,  Alex Manukyan 1, 0025 Yerevan, Republic of Armenia\\
              \email{rubengevorgyan@ysu.am}       
           \and
          	 Yenok B. Hakobyan \at
             Faculty of Economics and Management, Yerevan State University,  Alex Manukyan 1, 0025 Yerevan, Republic of Armenia\\
             \email{e.hakobyan@ysu.am}
}


\maketitle

\begin{abstract}
Cluster analysis is one of the essential tasks in data mining and knowledge discovery. Each type of data poses unique challenges in achieving relatively efficient partitioning of the data into homogeneous groups. While the algorithms for numeric data are relatively well studied in the literature, there are still challenges to address in case of categorical data. The main issue is the unordered structure of categorical data, which makes the implementation of the standard concepts of clustering algorithms difficult. For instance, the assessment of distance between objects, the selection of representatives for categorical data is not as straightforward as for numeric data. Therefore, this paper presents a new framework for partitioning categorical data, which does not use the distance measure as a key concept. The Matching based clustering algorithm is designed based on the similarity matrix and a framework for updating the latter using the feature importance criteria. The experimental results show this algorithm can serve as an alternative to existing ones and can be an efficient knowledge discovery tool.


\end{abstract}

\section{Introduction}
\label{intro}
Cluster analysis is one of the "super problems"s in data mining. Generally speaking, clustering is partitioning data points into intuitively similar groups (\citealt{SAXENA2017}).  This definition is simple and does not consider the challenges that occur while applying cluster analysis to real-world datasets. Nevertheless,  this type of analysis is common in different fields,e.g. text mining, marketing research, customer behavior analysis, financial market exploration. 
	
Nowadays various clustering algorithms have been developed in the literature. Each of them has its advantages and disadvantages. Moreover, as the data come in different forms, e.g.  text, numeric, categorical, image, the algorithms perform differently in different scenarios. In other words, the performance of a particular clustering algorithm depends on the structure of the data under consideration.

Cluster analysis of numeric data is relatively well studied in the literature. Various approaches are implemented such as representative-based, hierarchical,  density-based, graph-based, model-based, grid-based (\citealt{SAJANA2016}). Lately, increasing attention has been paid to partitioning non-numeric types of data. An important topic is the clustering of categorical data. The problem is that the most common clustering algorithms for categorical data are modifications of the ones introduced for numeric data. For instance, K-modes (\citealt{HUANG1997})  is a prototype of the K-means (\citealt{MACQUEEN1967}) algorithm. However, several researchers have developed algorithms specifically for categorical data (e.g. \citealt{NGUYEN2019254}, \citealt{CHEN2016322}, \citealt{YANTO201641}), but there is still much room for new approaches. 

The main problem in partitioning categorical data is that the implementation of the standard operations used in clustering algorithms possesses several limitations. For instance, the definition of distance between two objects with categorical features is not as straightforward as with numeric features, because categorical data takes only discrete values which do not have any order, unlike numeric data. The simplest solution is to transform the categorical data into binary data and then apply one of the common clustering algorithms. Moreover, several novel data transformation approaches have also been developed (e.g. \citealt{QIANLI201610}). On the other hand, researchers have developed and employed similarity measures (e.g. \citealt{BORIAH200810}, \citealt{GOUDA200510}, \citealt{Burnaby197003}) to overcome this issue. Another problem is the assessment of cluster representatives because many mathematical operations are not applicable to categorical data. For instance, it is impossible to assess the mean of the categorical feature. Taking into account the limitations of existing algorithms the aim of this paper is to present an algorithm which is not using predefined distance/similarity measures as a key concept and is not based on representatives for assigning data points to clusters. The key concept of the Matching based clustering algorithm ($MBC$)  is that two objects with categorical features are similar only if all the features match. Thus, the algorithm is based on the similarity matrix. In addition, we employ a feature importance framework for choosing which features to drop on each iteration until all objects are clustered. The test on the Soybean disease dataset (\citealt{DUA2017}) shows that the algorithm is highly accurate and can serve as an efficient data mining tool.

The rest of the paper is organized as follows.  We briefly review the common categorical data clustering algorithms in section \ref{sec:1}. In section \ref{sec:2} we discuss the categorical data and its limitations. In section \ref{sec:3} we introduce the general framework of the Matching based clustering algorithm. Section \ref{sec:4} presents the experimental results on the Soybean disease dataset. Finally, we summarize and describe our future plans.

\section{A review of categorical data clustering methods}
\label{sec:1}
Researchers have proposed various methods and algorithms for clustering categorical data. These algorithms can be grouped into five main classes: model-based, partition-based, density-base, hierarchical, projection-based (\citealt{Berkhin2006}). The main differences between these algorithms are the similarity or distance measures between data points and the criteria which identify the clusters. In the next paragraphs, we discuss the most common approaches in clustering categorical data.

Model-based clustering is based on the notion that data come from a mixture model. The most commonly used models are statistical distributions. This type of algorithms starts with assessing the prior model based on the user-specified parameters. Then the it aims at recovering the latent model by changing the parameters on each iteration. The main disadvantage of this type of clustering is that it requires user-specified parameters. Hence, if the assumptions are false, the results will be inaccurate. At the same time models may oversimplify the actual structure of the data. Another disadvantage of model-based clustering is that it can be slow on large datasets. Some model-based clustering algorithms are  SVM clustering (\citealt{WINTERS200702}), BILCOM Empirical Bayesian (\citealt{Andreopoulos2006}) ,Autoclass (\citealt{CHEESEMAN198854}). 

Partition-based clustering algorithms are the most commonly used ones because they are highly efficient in  processing large datasets. The main concept is defining representatives of each cluster, allocating objects to the cluster, redefining representatives and reassigning objects based on the dissimilarity measurements. This is repeated until the algorithm converges. The main drawback of this type of algorithms is that they require the number of clusters to be predefined by the user. Another disadvantage is that several algorithms of this type produce locally optimal solutions and are dependent on the structure of the dataset. Several  partition-based algorithms are K-modes, Fuzzy K-modes (\citealt{JiBAO201107}), Squeezer (\citealt{HEZENGYOU200209}), COOLCAT (\citealt{BARBARA2002}). 

Density-based algorithms define clusters as subspaces where the objects are dense and they are separated by subspaces of low density (e.g. \citealt{DUFANG2018}, \citealt{SINGH201710}, \citealt{AZZALINI2016}, \citealt{GIONIS2005}). The implementation of density-based algorithms for categorical data is challenging as the values of features are unordered. Even though they can be fast in clustering, they  may fail to cluster data with varying density.

Hierarchical algorithms represent the data as a tree of nodes, where each node is a possible grouping of data. There are two possible ways of clustering categorical data using hierarchical algorithms: in an agglomerative (bottom-up) and divisive (top-down) fashion. However, the latter is less common. The main concept of the agglomerative algorithm is using a similarity measure to gradually allocate the objects to the nodes of the tree. The main disadvantage of hierarchical clustering is their slow speed. Another problem is that the clusters may merge, thus these algorithms might lead to information distortion. Several  hierarchical clustering algorithms for categorical data are LIMBO (\citealt{ANDRITSOS2004}), ROCK (\citealt{GUHA2000}),  COBWEB (\citealt{Fisher1987}). 

Projected clustering algorithms are based on the fact that in high dimensional datasets clusters are formed based on specific subsets of features. In other words, each cluster is a subspace of high-dimensional datasets defined by a subset of features only relevant to that cluster. The main issue with projected clustering algorithms is that it requires user-specified parameters. If the defined parameters are inaccurate the clustering will be poor. Projected cluster algorithms include HIERDENC (\citealt{ANDREOPOULOS2007}), CLICKS (\citealt{ZAKIMARKUS2005}), STIRR (\citealt{GIBSON2000}), CLOPE (\citealt{YANGYILING2002}), CACTUS (\citealt{GANTI1999}).

Summarizing the existing algorithms we can conclude that most of them find some trade-off between accuracy and speed. However, considering the growing interest in analyzing categorical data in social, behavior, bio-medical science we are more interested in high accurate algorithms. Furthermore, as one can notice the majority of the algorithms uses some distance, similarity or density metrics and defines representatives of clusters as a subroutine of the algorithms. At the same time, they also require user-specified parameters. These factors can be seen as limitations in case of clustering categorical data. Therefore we propose a new approach to partitioning the categorical data, which  avoids these features. To introduce latter, in the next section we discuss the main characteristics of categorical data.

\section{Categorical data}
\label{sec:2}
Data comes in various forms such as numeric, categorical, mixture, spatial and so on. The analysis of each type of data poses unique challenges. The categorical data is not an exception. This type of data is widely used in political, social  and biomedical science. For instance, the measures of attitudes and opinions can be assessed with categorical data. The measures of the performance of medical treatments can also be categorical. Even though the mentioned fields have the largest influence on the development of the methods for categorical data, this type of data also occurs in other fields such as marketing, behavior science, education, psychology, public health, engineering. In this paper, we focus only on categorical features with unordered categories.

For sake of notation, consider a multidimensional dataset $D$ containing $n$ objects. Each object is described by $m$ categorical features each with $k = 1,2,3, ..$ unique categories. Thus, the dataset $D$ can be viewed as a matrix below:

\begin{equation}
D_{n,m} = 
 \begin{pmatrix}
  c_{1,1} & c_{1,2} & \cdots & c_{1,m} \\
  c_{2,1} & c_{2,2} & \cdots & c_{2,m} \\
  \vdots  & \vdots  & \ddots & \vdots  \\
  c_{n,1} & c_{n,2} & \cdots & c_{n,m} 
 \end{pmatrix}
\end{equation}

where each object is described by a set of categories $O_{i} = [c_{i,1}, c_{i,2}, c_{i,3} \cdots c_{i,m}]$. Also, the frequency of the unique category $c$ in the dataset is defined as $f(c)$.

As the categorical features have discrete values with no order, the application of distance measures such as euclidean distance may yield inaccurate results. However, the most common approach to overcome this limitation is the implementation of data transformation techniques. For instance, one can use binarization to transform the data into binary data and then apply the distance measures. On the other hand, the traditional way of comparing two objects with categorical features is to simply check if the categories coincide. If the categories of all the features under consideration match, the objects can be viewed as similar. This does not mean they are the same, because they can be  distinguished by other features. Thus, researchers have proposed various similarity measures instead of requiring all the features to match. The common approach is the overlap (\cite{STANFILL1986}). According to it, the similarity between two objects $O_x=[c_{x,1}, c_{x,2},  \cdots c_{x,m}]$ and $O_y=[c_{y,1}, c_{y,2}, \cdots c_{y,m}]$ is assessed by:

\begin{equation}
Ov(O_x,O_y) = \frac{1}{m}\sum_{i=1}^{m} \gamma_{i}, 
\end{equation}
where
\begin{equation}
  \gamma_{i} = 
  \begin{cases}
    1       & \quad \text{if } c_{x,i} = c_{y,i} \\
    0	  & \quad \text{otherwise }
  \end{cases}
\end{equation}

It can take values from $[0;1]$. The closer value gets to one, the higher is the similarity between the objects.

While implementing overlap,  one can notice that  the probability of finding two objects with the same categories rapidly decreases as the number of features and the number of unique categories of each feature increases. The problem is that the overlap measure gives equal weights to the features and doesn't take into account the importance of each feature in partitioning the data. However, the researchers have proposed more efficient ways of assessing similarity, which take into account the frequency of each category in the dataset. Researchers have introduced various types of similarity measures that are based on this concept (e.g. \citealt{JIAN2018}, \citealt{DOSSANTOS2015}, \citealt{AlAMURI2014}). Some of them are based on the probabilistic approaches, for instance, Goodall (\citealt{GOODALL1966}):

\begin{equation}
Goodall (O_x,O_y) = \frac{1}{m} \sum_{i=1}^{m} S_i, 
\end{equation}
where
\begin{equation}
 S_i = \begin{cases}
    1 - \frac{f( c_{x,i})(f( c_{x,i})-1)}{n(n-1)}       & \quad \text{if } c_{x,i} = c_{y,i} \\
    0	  & \quad \text{otherwise }
  \end{cases}
\end{equation}

Some are based on  information-theoretic approaches, for instance,  Lin (\citealt{LIN1998}).

\begin{equation}
Lin (O_x,O_y) =\frac{1}{\sum_{i=1}^m(log(\frac{f(c_{x,i})}{n})+ log(\frac{f(c_{y,i})}{n}))} \sum_{i=1}^{m} S_i, 
\end{equation}
where
\begin{equation}
 S_i = 
\begin{cases}
    2log(\frac{f(c_{x,i})}{n})       & \quad \text{if } c_{x,i} = c_{y,i} \\
    2(log(\frac{f(c_{x,i})}{n}+ \frac{f(c_{y,i})}{n}))	  & \quad \text{otherwise }
  \end{cases}
\end{equation}

Nevertheless, there are still cases when the use of similarity measures can be misleading. For instance, consider the $D_{4,2}$ dataset with 4 objects and  2 categorical features with $[a_1, a_2],[b_1, b_2]$ categories respectively.

\begin{table}[h]
\begin{center}
\label{tab:1}   
\caption {An example of data with 4 objects and 2 categorical features}
\begin{tabular}{cll}
\hline\noalign{\smallskip}
Object & C & B \\
\noalign{\smallskip}\hline\noalign{\smallskip}
$O_{1}$ & $a_{1}$ & $b_{1}$\\
$O_{2}$ & $a_{2}$ & $b_{2}$\\
$O_{3}$ & $a_{1}$ & $b_{2}$\\
$O_{4}$ & $a_{2}$ & $b_{1}$\\
\noalign{\smallskip}\hline
\end{tabular}
\end{center}
\end{table}

Based on this matrix the corresponding similarities between each unique pair of objects will be:

\begin{table}[h]
\begin{center}
\label{tab:1}   
\caption {The similarity measure between each unique pairs of objects in $D_{4,2}$}
\begin{tabular}{lccc}
\hline\noalign{\smallskip}
Object & Overlap & Lin & Goodall  \\
\noalign{\smallskip}\hline\noalign{\smallskip}
($O_{1},O_{2}$) & 0.00 & 0.00 & 0.00\\
($O_1,O_3$) & 0.50 & 0.50 & 0.42\\
($O_1,O_4$) & 0.50 & 0.50 & 0.42\\
($O_2,O_3$) & 0.50 & 0.50 & 0.42\\
($O_2,O_4$) & 0.50 & 0.50 & 0.42\\
($O_3,O_4$) & 0.00 & 0.00 & 0.00\\
\noalign{\smallskip}\hline
\end{tabular}
\end{center}
\end{table}

From the table, one can notice these measures can be misleading. For instance, one can group $O_{3}, O_{4}$ to either $O_{1}$ or $O_{2}$ as the similarity measures are the same. Therefore, similarity measures are powerful tools, but they should be used with caution. In this regard, one may consider using a quantitative measure to compare the features and choose relatively important ones. Then the objects will be similar if the categories of the selected features match. This is the main motivation of our approach.

Therefore, we employ several feature importance measures.  We define the partial grouping power ($PGP_I$) of the feature $l$  in dataset  $D$ as the number of unique matching pairs on the feature divided by the total number of  unique matching pairs in the dataset. This is based on the notion that if the feature has relatively higher number of matching pairs than others, it is more likely to group objects. The $PGP_I$  can be assessed by:
 
\begin{equation}
 PGP_l = \frac{\sum_{s=1}^{k_l}f(c_{s} )(f(c_{s} )- 1)}{\sum_{i=1}^{m} \sum_{j=1}^{k_i}f(c_{j} )(f(c_{j} )- 1)}
\end{equation}

where $c_{s}$ is the unique category of the feature, and $f(c_{s})$ is the frequency of the category in the dataset. This measure takes values from $[0;1]$. The closer the value to one the higher the importance of the feature in aggregating the objects.

We also define a measure for partitioning power of a feature. We define partial partitioning power ($PPP_l$) of the feature $l$  in dataset $D$ as the number of unique mismatching pairs on the feature divided by the total number of unique mismatching pairs in the dataset. The $PPP_l$ can be assessed by:

\begin{equation}
 PPP_l = \frac{n(n-1)-\sum_{s=1}^{k_l}f(c_{s} )(f(c_{s} )- 1)}{\sum_{i=1}^{m}( n(n-1)-\sum_{j=1}^{k_i}f(c_{j} )(f(c_{j} )- 1))}
\end{equation}

This measure takes values from $[0;1]$. The closer the value to one the higher the importance of the feature in partitioning the objects. Both methods can be used in the analysis. However, one of the measures can be more accurate than the other one depending on the data under consideration because the structure of the data may vary.

We also present another measure for assessing the feature importance. This one is based on the similarity matrix. Similarity matrix is defined as the matrix below:

\begin{equation}
SM_{n,n} = 
 \begin{pmatrix}
  m_{1,1} & m_{1,2} & \cdots & m_{1,m} \\
  m_{2,1} & m_{2,2} & \cdots & m_{2,m} \\
  \vdots  & \vdots  & \ddots & \vdots  \\
  m_{n,1} & m_{n,2} & \cdots & m_{n,m} 
 \end{pmatrix}
 \end{equation}

where $m_{i,j}$ = is a similarity measure between object $i$ and $j$ such as Overlap, Lin, Goodall. Through out this paper we will use the  count of matches ($CM$) between two objects as a similarity measure:

\begin{equation}
CM(O_x,O_y) = \sum_{i=1}^{m} \gamma_{i}, 
\end{equation}
where
\begin{equation}
 \gamma_{i} = 
  \begin{cases}
    1       & \quad \text{if } c_{x,i} = c_{y,i} \\
    0	  & \quad \text{otherwise }
  \end{cases}
\end{equation}

The similarity matrix is symmetrical, thus only  upper triangular matrix is used in the calculations. Furthermore, the diagonal will also be ignored. Based on the similarity matrix we define the general influence matrix($GIM$)  as :

\begin{equation}
IM_{n,n} = 
 \begin{pmatrix}
  I_{1,1} & I_{1,2} & \cdots & I_{1,m} \\
  I_{2,1} & I_{2,2} & \cdots & I_{2,m} \\
  \vdots  & \vdots  & \ddots & \vdots  \\
  I_{n,1} & I_{n,2} & \cdots & I_{n,m} 
 \end{pmatrix}
\end{equation}
where
\begin{equation}
 I_{i,j} = 
\begin{cases}
   1      & \quad \text{if } m_{i,j} > \alpha \\
   0  & \quad \text{otherwise }
  \end{cases}
\end{equation}

where $\alpha$ is a threshold, which is bounded by the values similarity measure can take. In this paper, we set $\alpha$ to 0. After the construction, the features or the subset of features under consideration are dropped, and the influence matrix is updated. The matrix  after the drop is defined as the partial influence matrix ($PIM_l$) of corresponding feature or subset of features $l$. In this case, the partial grouping power ($PGP2_l$) of the feature or subset of features  $l$, is assessed by dividing the count of the ones in the $PIM_l$  by the count of ones in the $GIM$.

\begin{equation} 
 PGP2_l = \frac{\eta_{PIM_l}}{\eta_{GIM}}, 
\end{equation}

where $ \eta_{PIM_l}$ is the count of ones in the $PIM_l$ and  $\eta_{GIM}$ is the count of ones in the $GIM$. 

One can notice that these measures of feature importance depend only on the number of unique matches in the dataset, and the number of categories of each feature does not influence them. In the next section, we present the Matching based clustering ($MBC$) algorithm, which combines  the importance measures of the features and the similarity matrix to partition categorical data into homogeneous groups.

\section{Matching based clustering algorithm}
\label{sec:3}
Similar to any clustering algorithm the main objective of MBC is partitioning the data into relatively similar groups. The algorithm is defined for categorical data only. However, one can modify it to employ for other types also, but this is out of the scope of this paper. The main idea is, while there are still objects without clusters, the algorithm will choose features to drop  based on their importance. Then it will update the similarity matrix and try to cluster the objects based on the new $SM$. It uses the similarity matrix where  the similarity measure between two objects is defined by the $CM$. We also use either $PGP_I$ or $PPP_I$ measure to choose the features to drop on each iteration. For the sake  of notations, we define $\theta_p$ as the count of the remaining features on iteration $p$. The initial value of $\theta_0 = m$. We consider  two objects to belong to the same cluster if $ m_{i,j} = \theta_p$. In other words, they are grouped if their categories coincide for all the remaining features on iteration  $p$.

The algorithm consists of the following steps:
\begin{enumerate}
\item Construct the similarity matrix $SM_{n,n}$.
\item Calculate the $PGP_I$ of each feature.
\item Allocate the objects to clusters based on the similarity matrix. In other words, group two objects ($i$ and $j$), if $ m_{i,j} = \theta_p$. If one of the objects  is already allocated to a cluster, assign the second one to the same cluster. 
\item Check if there are still objects not assigned to any cluster, if yes continue to next step, otherwise terminate.
\item Remove the features with the lowest $PGP_I$. If there are several features, one may consider either dropping all of them or using the $PGP2_I$ to choose which one to drop.
\item Update the similarity matrix.
\item Additionally update the $SM$ using the statement: $\forall$ existing clusters  $i$ and $j$, if  $m_{i,j} =  \theta_p$, then the values of the rows and columns $i$ and $j$, which  are equal to $\theta_p$, are set to zero.
\item  Return to step 3.
\end{enumerate}

The algorithm stops if all the objects are clustered or the importance  of remaining features is the same. As one can notice the algorithm can use also the $PPP_I$  as feature importance measure. In this case, the features with the lowest values should be dropped. Moreover, step 7 is optional. The main purpose of this step is to avoid the merging of existing clusters,  if one avoids this step, the algorithm builds the dendrogram of the data, where each level of the tree corresponds to the clusters after each feature drop. 

\begin{figure}[h]
\begin{center}
\includegraphics[scale=0.1]{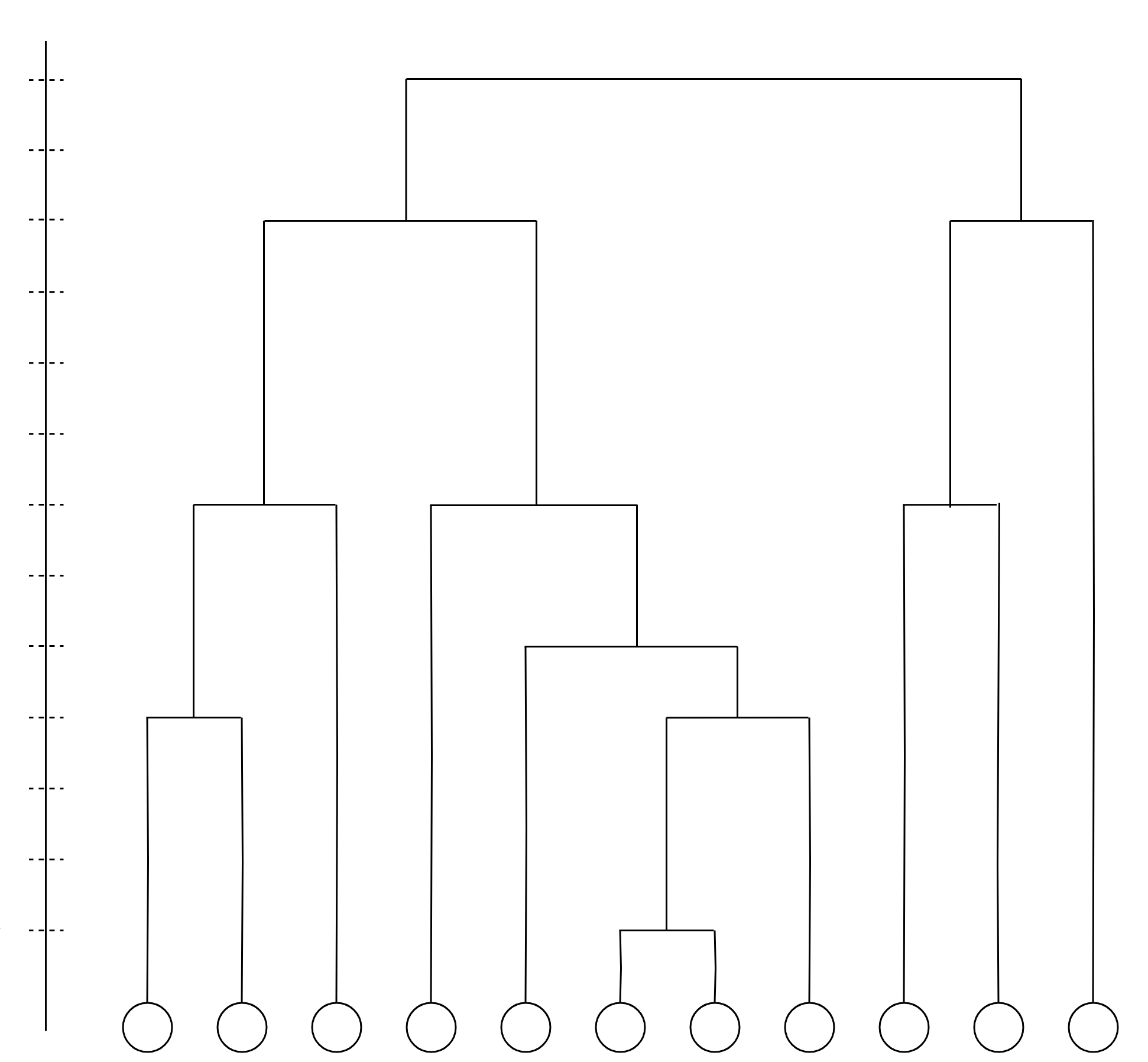}
\end{center}
\caption {An example of the dendrogram build by $MBC$ algorithm}
\end{figure}

To illustrate how the algorithm works, we will apply it to a simulated dataset $A$.

\begin{table}[h]
\label{tab:2} 
\begin{center}
\caption {Simulated dataset $A$}
\begin{tabular}{lllllll}
\hline\noalign{\smallskip}
Objects&A&B&C&D&E  \\
\noalign{\smallskip}\hline\noalign{\smallskip}
$O_1$&$a_2$&$b_1$&$c_2$&$d_3$&$e_2$\\
$O_2$&$a_2$&$b_1$&$c_2$&$d_3$&$e_2$\\
$O_3$&$a_2$&$b_1$&$c_2$&$d_3$&$e_1$\\
$O_4$&$a_2$&$b_1$&$c_2$&$d_3$&$e_4$\\
$O_5$&$a_1$&$b_2$&$c_4$&$d_2$&$e_3$\\
$O_6$&$a_1$&$b_2$&$c_3$&$d_4$&$e_4$\\
$O_7$&$a_1$&$b_2$&$c_4$&$d_2$&$e_2$\\
$O_8$&$a_1$&$b_2$&$c_3$&$d_4$&$e_1$\\
$O_9$&$a_1$&$b_2$&$c_1$&$d_1$&$e_3$\\
$O_{10}$ &$a_1$&$b_2$&$c_4$&$d_2$&$e_2$\\
\noalign{\smallskip}\hline
\end{tabular}
\end{center}
\end{table}

In this dataset 10 objects are defined by 4 categorical features $A, B, C, D, E $ with $[a_1, a_2]$, $[b_1, b_2]$, $[c_1, c_2, c_3, c_4]$, $[d_1, d_2, d_3, d_4]$ and $[e_1, e_2, e_3, e_4]$ unique categories respectively. 

We initialize the algorithm by constructing the similarity matrix:

\begin{equation}
S_{10,10} =
\begin{pmatrix}
-&5&4&4&0&0&1&0&0&1\\
-&-&4&4&0&0&1&0&0&1\\
-&-&-&4&0&0&0&1&0&0\\
-&-&-&-&0&1&0&0&0&0\\
-&-&-&-&-&2&4&2&3&4\\
-&-&-&-&-&-&2&4&2&2\\
-&-&-&-&-&-&-&2&2&5\\
-&-&-&-&-&-&-&-&2&2\\
-&-&-&-&-&-&-&-&-&2\\
-&-&-&-&-&-&-&-&-&-
\end{pmatrix}
\end{equation}

Then, the importance of each feature is assessed. In this example, we will use the $PGP_I$ measure. For instance the $PGP_A$ will be:

\begin{equation}
PGP_A = \frac{21}{21+21+10+10+9} = 0.30
\end{equation}

Respectively $PGP_B=0.30,  PGP_C=0.14, PGP_D=0.14$ and  $PGP_E=0.13$. Then as the $\theta_0 = m = 5$, all the objects $i,j$ with $m_{i,j} = 5$  are grouped. As we can see we have two clusters $[O_1,O_2]$ and $[O_7,O_{10}]$. As we still have some objects left without cluster, we continue to next step. In particular, as the feature ${E}$ has the lowest $PGP$ , we drop it and update the similarity matrix. Also to avoid the merging of already existing clusters,  we additionally update $SM$ according to step 7. As similarity between clusters $[O_1,O_2]$ and $[O_7,O_{10}]$ is not equal to $\theta_1 = 4$, we will not make additional changes, and the new  data view and corresponding similarity matrix will be: 

\begin{center}
\[
\begin{array}{llllll}
\hline\noalign{\smallskip}
Objects&A&B&C&D  \\
\noalign{\smallskip}\hline\noalign{\smallskip}
(O_1,O_2)&a_1&b_0&c_1&d_2\\
O_3&a_1&b_0&c_1&d_2\\
O_4&a_1&b_0&c_1&d_2\\
O_5&a_0&b_1&c_3&d_1\\
O_6&a_0&b_1&c_2&d_3\\
(O_7,O_{10})&a_0&b_1&c_3&d_1\\
O_8&a_0&b_1&c_2&d_3\\
O_9&a_0&b_1&c_0&d_0\\
\noalign{\smallskip}\hline
\end{array}
\quad	
S_{8,8} =
\begin{pmatrix}
-&4&4&0&0&0&0&0\\
-&-&4&0&0&0&0&0\\
-&-&-&0&0&0&0&0\\
-&-&-&-&2&4&2&2\\
-&-&-&-&-&2&4&2\\
-&-&-&-&-&-&2&2\\
-&-&-&-&-&-&-&2\\
-&-&-&-&-&-&-&-

\end{pmatrix}
\]
\captionof{figure}{The data view and corresponding $SM$ after the first iteration}
\end{center}

As $\theta_1 = 4$, we will have $[O_1,O_2,O_3,O_4]$, $[O_5,O_7,O_{10}]$ and $[O_6,O_8]$ clusters. However, we still have one more object to assign to a cluster, thus we drop $C$ and $D$ and update the similarity matrix.

\begin{center}
\[
\begin{array}{llllll}
\hline\noalign{\smallskip}
Objects&A&B \\
\noalign{\smallskip}\hline\noalign{\smallskip}
(O_1,O_2,O_3,O_4)&a_1&b_0\\
(O_5,O_7,O_{10})&a_0&b_1\\
(O_6,O_8)&a_0&b_1\\
O_9&a_0&b_1\\
\noalign{\smallskip}\hline
\end{array}
\quad	
S_{4,4} = 
\begin{pmatrix}
-&0&0&0\\
-&-&2&2\\
-&-&-&2\\
-&-&-&-
\end{pmatrix}
\]
\captionof{figure}{The data view and corresponding $SM$ after the second iteration}
\end{center}

But we also check the statement of step 7 between any pair of  the existing clusters. As  $\theta_2 = 2$,  the statement is true in the case of  $[O_5,O_7,O_{10}]$ and $[O_6,O_8]$. Thus, the values of corresponding rows(1,3) and columns(1,3), which are equal  $\theta_2 = 2$, are set to zero. The purpose of this modification is that as we are dropping features with low grouping power, the clusters are more likely to merge. Therefore, we may lose important local partitioning of data points. Thus, the final updated similarity matrix will be :

\begin{equation}
S_{4,4} = 
\begin{pmatrix}
-&0&0&0\\
-&-&0&0\\
-&-&-&0\\
-&-&-&-
\end{pmatrix}
\end{equation}

However, the second iteration does not group the $O_9$. At the same time as the importance of the remaining features $A$ and $B$ is the same the algorithm terminates and the object $O_9$ forms the fourth cluster. Thus, the final result of the MBC clustering is :

\begin{table}[h]
\begin{center}
\caption {The final clusters for dataset $A$}
\label{tab:1}       

\begin{tabular}{llllllc}
\hline\noalign{\smallskip}
Objects&A&B&C&D&E&Cluster  \\
\noalign{\smallskip}\hline\noalign{\smallskip}
$O_1$&$a_1$&$b_0$&$c_1$&$d_2$&$e_2$&1\\
$O_2$&$a_1$&$b_0$&$c_1$&$d_2$&$e_2$&1\\
$O_3$&$a_1$&$b_0$&$c_1$&$d_2$&$e_1$&1\\
$O_4$&$a_1$&$b_0$&$c_1$&$d_2$&$e_4$&1\\
$O_5$&$a_0$&$b_1$&$c_3$&$d_1$&$e_3$&2\\
$O_6$&$a_0$&$b_1$&$c_2$&$d_3$&$e_4$&3\\
$O_7$&$a_0$&$b_1$&$c_3$&$d_1$&$e_2$&2\\
$O_8$&$a_0$&$b_1$&$c_2$&$d_3$&$e_1$&3\\
$O_9$&$a_0$&$b_1$&$c_0$&$d_0$&$e_3$&4\\
$O_{10}$&$a_0$&$b_1$&$c_3$&$d_1$&$e_2$&2\\
\end{tabular}
\end{center}
\end{table}

The algorithm has some unique characteristics worth mentioning. First, to achieve better performance one can notice that all the changes required in each step should be done only on the similarity matrix and there is no need to update the dataset. Second, there is no need for user-defined parameters. However, one may specify the number of clusters. In this case, step 7 is ignored and after the construction of the dendrogram, the clusters are formed. The algorithm  creates a tree where each leaf is a possible cluster. In the case of the simulated dataset $A$,  the dendrogram will be:

\begin{figure}[h]
\begin{center}
\includegraphics[scale=0.1]{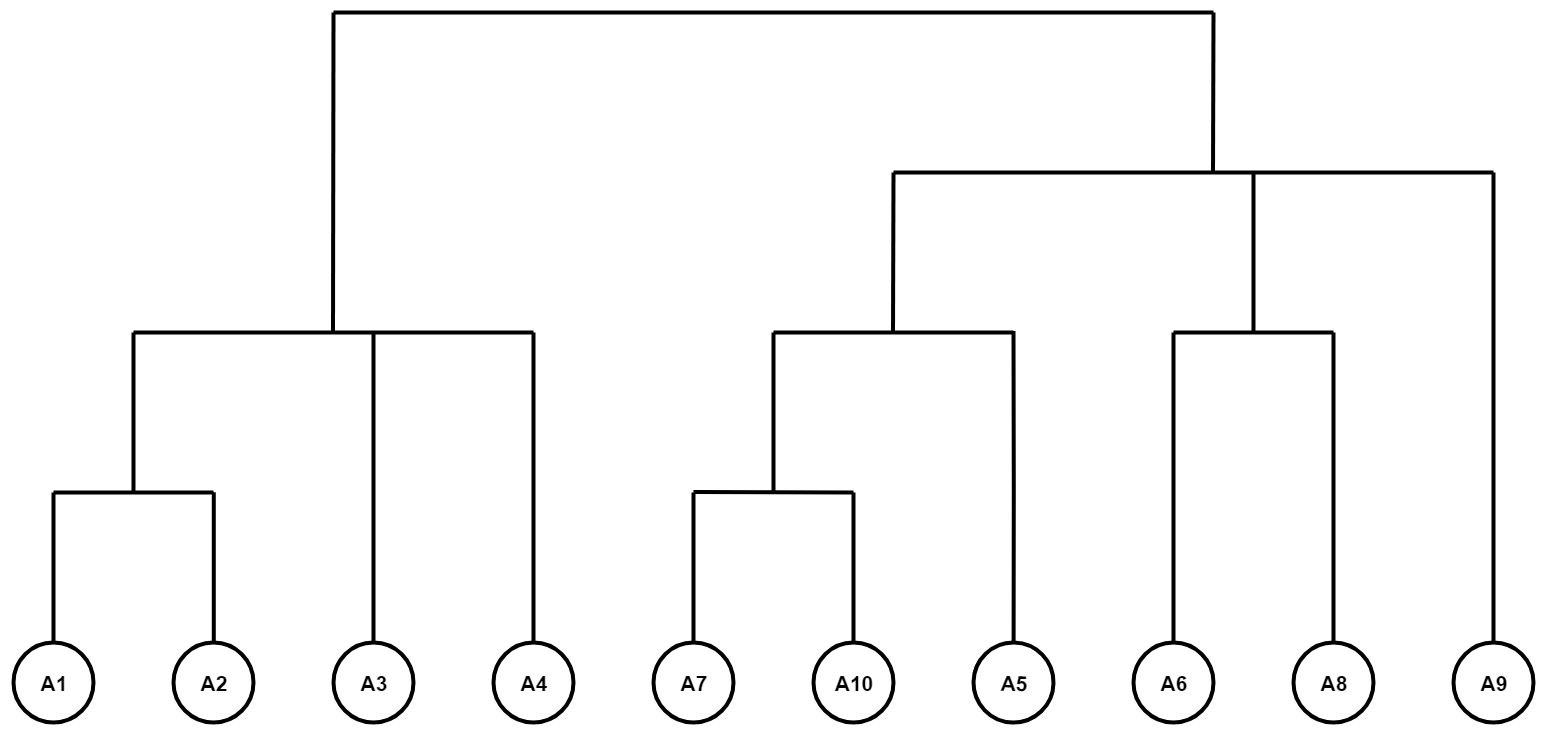}
\end{center}
\caption {The dendrogram of clusters of the simulated dataset $A$}
\end{figure}

Forth, as the algorithm is based on either feature grouping or partitioning power, this information can be used to understand the structure of the data. For instance, this algorithm can serve as a subroutine for feature selection for other clustering algorithms.

\section{Experimental results}
\label{sec:4}
We have employed MBC to the Soybean disease dataset (\citealt{DUA2017}) to test its performance on the real-world dataset.  It is one of the standard test data sets used in the machine learning community and has often been used to test conceptual clustering algorithms for categorical data. The Soybean data set has 47 observations, each being described by 35 features. Each observation is identified by one of the 4 diseases -- Diaporthe Stem Canker, Charcoal Rot, Rhizoctonia Root Rot, and Phytophthora Rot. These are used as indicators of the accuracy of the algorithm. 

After applying the $MBC$ to the Soybean disease dataset, we got 18 different clusters.

\begin{table}[h]
\caption {The distribution of clusters by 4 types of diseases}
\begin{center}
\begin{adjustbox}{width=\textwidth*45/100}
\begin{tabular}{lllllllllllllllllll}
\hline\noalign{\smallskip}
DT&1&2&3&4&5&6&7&8&9&10&11&12&13&14&15&16&17&18\\
\noalign{\smallskip}\hline\noalign{\smallskip}
D1& -   &3&2& -   & -   & -   & -   & -   & -   &2&2& -   & -   & -   & -   & -   & -   &1\\
D2& -   & -   & -   &3& -   & -   &2&2&3& -   & -   & -   & -   & -   & -   & -   & -   & -   \\
D3& -   & -   & -   & -   &2& -   & -   & -   & -   & -   & -   &5&2& -   & -   & -   & -   &1\\
D4&5& -   & -   & -   & -   &3& -   & -   & -   & -   & -   & -   & -   &2&3&2&2& -   \\
\end{tabular}
\end{adjustbox}
\end{center}
\end{table}

As we can see from the table above all the clusters except for one entirely belong to one of the groups mentioned above.  In other words,  we have only one possible misclassification. However, as already mentioned one may require a specific number of clusters. In this case, one can use the constructed dendrogram based on the MBC algorithm.

\begin{figure}[h]
\begin{center}
\includegraphics[scale=0.04]{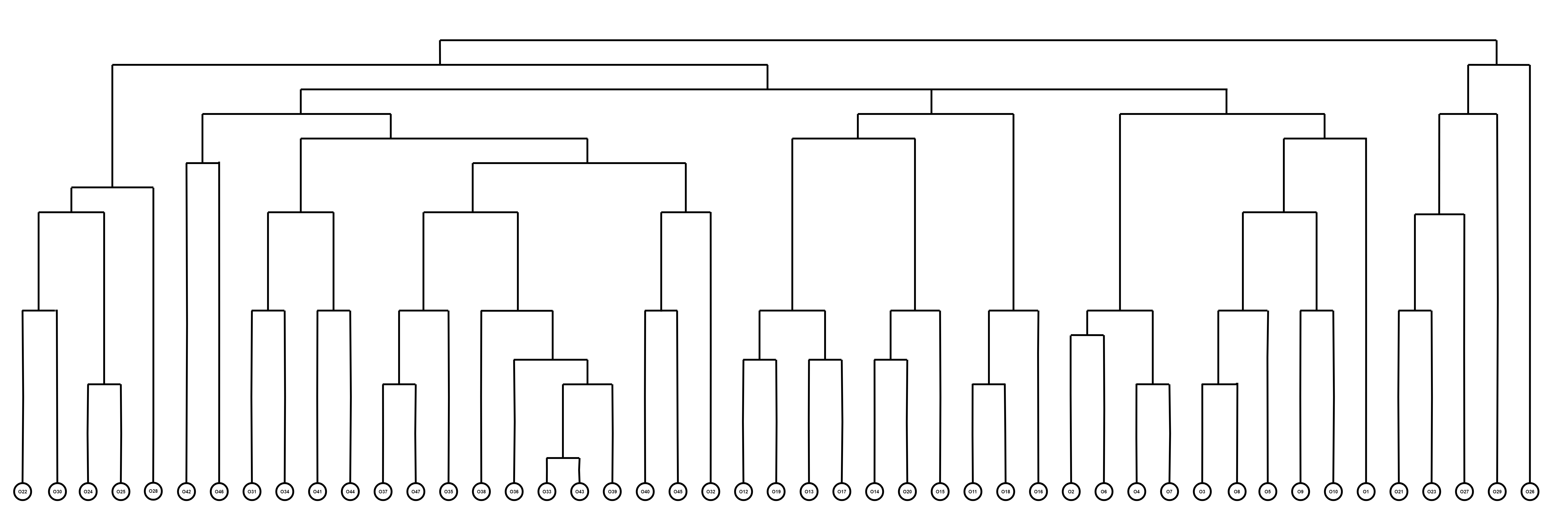}
\end{center}
\caption {The dendrogram for the Soybean disease dataset}
\end{figure}

If we compare the MBC with K-modes (\citealt{HUANG1997}), the main differences are: K-modes depends on the data order and the initial cluster representatives, it also requires the number of clusters as an input parameter. In the case of MBC, we do not have these limitations. 

\section*{Conclusion}
\label{conc}
Data clustering is broadly studied and used in various applications. The best practice approaches are limited to numeric data, but lately, increasing attention has been paid to clustering other types of data, in particular, categorical data. Thus, specific models are being developed for categorical data. The vital issue in clustering  categorical data is the notion of the distance or similarity between the observations. Hence, this paper introduces the Matching based clustering algorithm which can serve as an alternative to existing ones. It is based on the main characteristics of categorical data and presents a new framework for clustering categorical data, which is not based on the distance measure.  The main concept of the algorithm is the assessment of the similarity matrix, updating the  latter based on the importance criteria of each feature or subset of features and grouping only if the categories of objects entirely match. These operations allow clustering categorical data without transformation. Another advantage is the description of the features, as the algorithm allows to identify the ones which cause the partitioning of the data. These can be important in interpreting clustering results. Finally, MBC does not require any initial parameter.

Our future work plan is to develop and implement a modification of the algorithm to cluster mixture data. Furthermore, overcome its limitation and adapt it to clustering big datasets.  Such an algorithm is required in a number of data mining applications, such as partitioning large  sets of objects into a number of smaller and more manageable subsets that can be more easily modeled and analyzed.


\bibliography{reference}

\end{document}